\documentclass[conference]{IEEEtran}

\IEEEoverridecommandlockouts

\usepackage{amsmath,amssymb,amsfonts,amsthm}
\usepackage{algorithmic}
\usepackage{graphicx}
\usepackage{xcolor}
\usepackage{textcomp}
\usepackage{multirow}
\usepackage{float}
\usepackage[caption=false,font=footnotesize]{subfig} % IEEEtran-recommended
\usepackage{overpic}
\usepackage{placeins}
\usepackage{stfloats}     % For bottom placement of two-column figures
\usepackage{cite}
\usepackage[hidelinks]{hyperref}

\hypersetup{
  pdfinfo={
    Title={Kolmogorov--Arnold Representation for Symplectic Learning: Advancing Hamiltonian Neural Networks},
    Author={Zongyu Wu, Ruichen Xu, Luoyao Chen, Georgios Kementzidis, Siyao Wang, Yuefan Deng},
    Subject={Accepted to IEEE IJCNN; arXiv version},
    Keywords={Kolmogorov--Arnold Networks, Hamiltonian Neural Networks, Symplectic Structure, Energy Conservation}
  }
}

\newtheorem{theorem}{Theorem}

\def\BibTeX{{\rm B\kern-.05em{\sc i\kern-.025em b}\kern-.08em
    T\kern-.1667em\lower.7ex\hbox{E}\kern-.125emX}}

\begin{document}

\title{Kolmogorov--Arnold Representation for Symplectic Learning: Advancing Hamiltonian Neural Networks}

\author{
\IEEEauthorblockN{1\textsuperscript{st} Zongyu Wu\textsuperscript{*}}
\IEEEauthorblockA{\textit{Independent Researcher}\\
United States\\
\texttt{zongyuwu97@gmail.com}}
\and
\IEEEauthorblockN{2\textsuperscript{nd} Ruichen Xu\textsuperscript{*}}
\IEEEauthorblockA{\textit{Stony Brook University}\\
Stony Brook, NY, United States\\
\texttt{ruichen.xu@stonybrook.edu}}
\and
\IEEEauthorblockN{3\textsuperscript{rd} Luoyao Chen}
\IEEEauthorblockA{\textit{New York University}\\
New York, NY, United States\\
\texttt{lc4866@nyu.edu}}
\and
\IEEEauthorblockN{4\textsuperscript{th} Georgios Kementzidis}
\IEEEauthorblockA{\textit{Stony Brook University}\\
Stony Brook, NY, United States\\
\texttt{georgios.kementzidis@stonybrook.edu}}
\and
\IEEEauthorblockN{5\textsuperscript{th} Siyao Wang}
\IEEEauthorblockA{\textit{University of California, Davis}\\
Davis, CA, United States\\
\texttt{syywang@ucdavis.edu}}
\and
\IEEEauthorblockN{6\textsuperscript{th} Yuefan Deng\textsuperscript{\dag}}
\IEEEauthorblockA{\textit{Stony Brook University}\\
Stony Brook, NY, United States\\
\texttt{yuefan.deng@stonybrook.edu}}
\thanks{\textsuperscript{*}Zongyu Wu and Ruichen Xu contributed equally to this work.}
\thanks{\textsuperscript{\dag}Corresponding author.}
}

\IEEEaftertitletext{%
\vspace{-0.5ex}\noindent\begin{minipage}{\textwidth}\centering\footnotesize
This work has been accepted by IJCNN 2025; a formal version will appear in the IEEE/IJCNN proceedings.\\
This is the arXiv v1 submission (2025-08-26).
\end{minipage}\par\medskip}

\maketitle
\begin{abstract}
%\ZW{removed duplicate section `Evaluation`, revised model evaluation methods to reflect actual evaluation.}

%\LC{rewrite this as: Data-driven modeling of physical systems has advanced at the intersection of scientific computing and machine learning, particularly for problems with partially unknown or high-dimensional dynamics. While Hamiltonian Neural Networks (HNNs) ensure energy conservation by learning Hamiltonian functions from data, existing implementations often rely on Multilayer Perceptrons (MLPs), which are sensitive to hyperparameters and struggle with complex energy landscapes.
%In this work, we propose the Kolmogorov–Arnold Representation Theorem-based Hamiltonian Neural Network (KAR-HNN), which replaces the MLP with univariate transformations inspired by the Kolmogorov–Arnold theorem. This approach leverages localized function approximations to better capture high-frequency and multi-scale dynamics, reducing energy drift and improving long-term predictive stability. The network structure also preserves the symplectic form of Hamiltonian systems, maintaining interpretability and physical consistency.
%We validate KAR-HNN on standard benchmarks, demonstrating its effectiveness for accurate and stable modeling of real-world physical processes.}

We propose a Kolmogorov–Arnold Representation-based Hamiltonian Neural Network (KAR-HNN) that replaces the Multilayer Perceptrons (MLPs) with univariate transformations. While Hamiltonian Neural Networks (HNNs) ensure energy conservation by learning Hamiltonian functions directly from data, existing implementations, often relying on MLPs, cause hypersensitivity to the hyperparameters while exploring complex energy landscapes.
Our approach exploits the localized function approximations to better capture high-frequency and multi-scale dynamics, reducing energy drift and improving long-term predictive stability. The networks preserve the symplectic form of Hamiltonian systems, and, thus, maintain interpretability and physical consistency. After assessing KAR-HNN on four benchmark problems including spring-mass, simple pendulum, two- and three-body problem, we foresee its effectiveness for accurate and stable modeling of realistic physical processes often at high dimensions and with few known parameters.

\end{abstract}


\section{Introduction}
Modeling high-fidelity dynamics in physical systems remains a central challenge in both scientific computing and artificial intelligence. Classical approaches, such as symplectic integrators \cite{Hairer2006}, presuppose explicit knowledge of the underlying physics, relying on specialized numerical schemes to preserve geometric properties. Being effective under well-understood conditions, these methods become difficult for problems where the exact governing equations are unknown or only partially known, particularly in high-dimensional or partially observed settings \cite{Celledoni2021,Azencot2022, xu2025velocityinferredhamiltonianneuralnetworks}. Consequently, it is challenging to adapt purely classical methods to learn the governing laws directly from data. In contrast, recent advances at the intersection of machine learning and physics offer \emph{data-driven} techniques \cite{Chen2018,Raissi2019} that infer system dynamics from observations while still maintaining essential physical constraints.

%Among these advances, \emph{Hamiltonian Neural Networks (HNNs)} have emerged as a framework for learning a Hamiltonian function \(H_\theta\) whose partial derivatives generate energy-conserving dynamics \cite{Greydanus2019}. \LC{what does this mean? While...} While HNNs typically adopt Multilayer Perceptrons (MLPs) to represent \(H_\theta\), MLPs can require painstaking hyperparameter tuning to capture complex or highly localized features in a system’s energy landscape \cite{Werbos1982,Chang2022,Song2022,Wang2023,Lee2023}. This shortcoming can lead to inaccuracies over long time horizons and diminished interpretability. \LC{rewrite: Typically, \(H_\theta\) is modeled using Multilayer Perceptrons (MLPs), which requires hyperparameter tuning to capture complex or localized features in a system's energy landscape. This can hinder interpretability and, if parameters are not optimized, lead to reduced accuracy over long time horizons.}
Among these advances, \emph{Hamiltonian Neural Networks (HNNs)} have emerged as a framework for learning a Hamiltonian function \(H_\theta\) whose partial derivatives generate energy-conserving dynamics \cite{Greydanus2019}. Typically, \(H_\theta\) is modeled using Multilayer Perceptrons (MLPs), which require hyperparameter tuning to capture complex or localized features in a system’s energy landscape. This approach can hinder interpretability and, if parameters are not carefully optimized, lead to reduced accuracy over long time horizons.

% \noindent \textbf{Kolmogorov--Arnold Network (KAN).}
% On the other hand, researchers have explored \emph{Kolmogorov--Arnold Networks (KANs)} to improve function approximation by leveraging the Kolmogorov--Arnold theorem \cite{Kolmogorov1957,Arnold1957,Pan2019,Mishra2022,ZimingLiu2023}. Instead of relying on fully connected layers as in MLPs, \LC{Unlike MLPs,}KANs decompose an \(n\)-variate function into sums and compositions of univariate transformations. This localized architecture can more effectively capture high-frequency or oscillatory features and often leads to smoother gradient fields. Empirically, KANs have demonstrated better data efficiency and interpretability, especially when modeling multi-scale physical phenomena \cite{Tang2020}. For example, Liu et al.\ \cite{ZimingLiu2023} showed that exploiting the Kolmogorov--Arnold superposition principle can significantly reduce the complexity required to approximate intricate, high-dimensional functions, thus underscoring the potential of KAN designs for physics-based modeling tasks.

On the other hand, researchers have explored \emph{Kolmogorov--Arnold Networks (KANs)} to improve function approximation by leveraging the Kolmogorov–Arnold Representation \cite{Kolmogorov1957,Arnold1957,Pan2019,Mishra2022,ZimingLiu2023}. \textbf{Unlike MLPs,} KANs decompose an \(n\)-variate function into sums and compositions of univariate transformations, forming a localized architecture that can more effectively capture high-frequency or oscillatory features while often yielding smoother gradient fields. Empirically, KANs have demonstrated better data efficiency and interpretability when modeling multi-scale physical phenomena \cite{Tang2020}. For example, Liu et al.\ \cite{ZimingLiu2023} showed that exploiting the Kolmogorov–Arnold Representation can significantly reduce the complexity required to approximate intricate, high-dimensional functions, thereby underscoring the potential of KAN designs for physics-based modeling tasks.

% \noindent \textbf{Kolmogorov--Arnold Representation Theorem-based Hamiltonian Neural Network (KAR-HNN).} 
%Building on these insights, we propose a novel architecture: \emph{the Kolmogorov–Arnold Representation Theorem-Based Hamiltonian Neural Network (KAR-HNN)}. This approach replaces the MLP in HNNs with a Kolmogorov--Arnold decomposition \cite{Kolmogorov1957,Arnold1957}. This theorem \LC{which theorem?If this refer to KAR-HNN, then use "which"} establishes that any continuous multivariate function can be expressed as compositions and sums of univariate transformations, thereby providing a constructive method for tackling high-dimensional Hamiltonians. By adopting a KAN-like, univariate-based approach, KAR-HNN inherits \textbf{two key advantages}:

Building on these insights, we propose a novel architecture: \emph{the Kolmogorov--Arnold Representation-based Hamiltonian Neural Network (KAR-HNN)}. This approach replaces the MLP in HNNs with a Kolmogorov–Arnold Representation \cite{Kolmogorov1957,Arnold1957}, which states that any continuous multivariate function can be expressed as compositions and sums of univariate transformations, thereby providing a constructive method for tackling high-dimensional Hamiltonians. By adopting a KAN-like, univariate-based approach, KAR-HNN inherits \textbf{two key advantages}:

\begin{enumerate}
    \item \textbf{Localized Function Approximation:} Similar to KAN, KAR-HNN leverages sums of univariate subfunctions to handle sharp gradients, oscillatory regions, or multi-scale energy landscapes. This approach often reduces the need for extensive hyperparameter tuning compared to fully connected MLPs.
    \item \textbf{Preservation of Symplectic Structure:} Crucially, the KAR-HNN keeps the canonical form of Hamilton’s equations,
    \[
    \dot{q} = \frac{\partial H_\theta}{\partial p}, \qquad
    \dot{p} = -\,\frac{\partial H_\theta}{\partial q},
    \]
    ensuring that the learned dynamics remain symplectic. That is, energy conservation and phase-space geometry are preserved in the same way as classical Hamiltonian systems \cite{Marsden1999, SanzSerna1994}. As a result, KAR-HNN exhibits smaller energy drift over long time horizons and maintains more faithful trajectories.
\end{enumerate}

We demonstrate these benefits across representative benchmark problems, including pendulum systems, spring-mass oscillators, and multi-body orbital dynamics. 

The rest of this paper is organized as follows. Section \ref{sec:pre} outlines the fundamentals of Hamiltonian dynamics and standard HNN frameworks. Section \ref{sec:method} introduces how to construct our KAR-HNN, including how the Kolmogorov–Arnold Representation is implemented. We then present experimental results in Section \ref{sec:Experiment}, showing that our method consistently outperforms MLP-based baselines in accuracy, robustness, and physical fidelity. %Lastly, we discuss future directions and potential applications of the Kolmogorov--Arnold theorem in broader physics-informed learning contexts in Section \ref{sec:ExperimentEvaluation}. \LC{Missing section for future direction} 

%\LC{recommend including  "limitations" into the last section too.}

\noindent \textbf{Contributions:}
\begin{itemize}
    \item We propose a novel \emph{KAR-HNN} architecture, leveraging the Kolmogorov--Arnold representation to approximate the Hamiltonian function, enabling smoother gradient fields and better handling of localized or oscillatory phenomena.
    \item We provide theoretical insights showing why decomposing into univariate functions can improve the approximation of Hamiltonian systems, especially in high-dimensional scenarios, while preserving the canonical symplectic form.
    \item We empirically validate our method on benchmark problems (e.g., pendulums, multi-body orbits), demonstrating higher accuracy, reduced energy drift, and robustness compared to MLP-based HNNs.
\end{itemize}
\begin{figure*}[t]
    \centering
    \includegraphics[width=\textwidth]{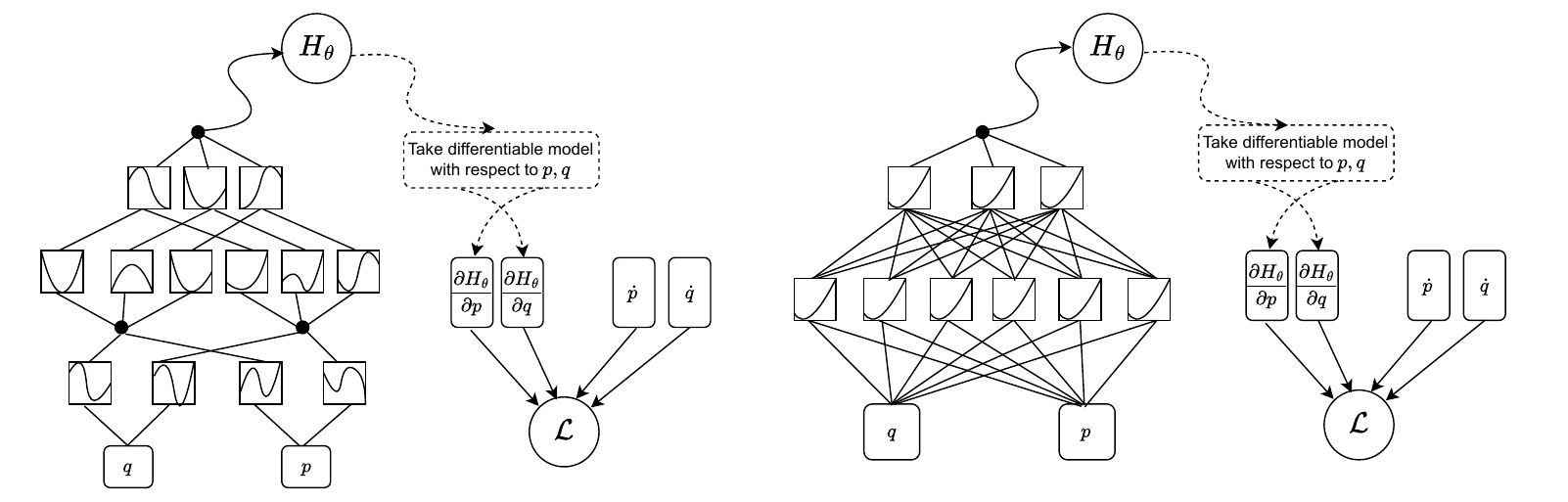}
    \caption{\textbf{The juxtaposition of Kolmogorov–Arnold Representation and MLP-Based Hamiltonian Neural Networks I would like to check whether it is the good name of the caption of the graph.}
    The left panel (KAN-HNN) models the Hamiltonian \(H_\theta\) as a superposition of univariate “kernel” or “basis” blocks (shown as small plots), following the Kolmogorov–Arnold Representation. Each block contributes to the final energy function in a localized manner. The right panel (MLP-HNN) uses a fully connected feedforward network to represent \(H_\theta\). In both cases, partial derivatives \(\tfrac{\partial H_\theta}{\partial p}\) and \(\tfrac{\partial H_\theta}{\partial q}\) are computed and compared against observed \(\dot{p}\) and \(\dot{q}\) through a differentiable loss \(\mathcal{L}\). Automatic differentiation handles the chain rule with respect to the learnable parameters \(\theta\), enabling end-to-end training that enforces Hamilton’s equations.}
    \label{fig:hnn}
\end{figure*}
\section{Background}
\label{sec:RelatedWork}

Over the past decade, synergy between machine learning (ML) and dynamical systems theory has substantially shifted how researchers tackle complex physical processes. By uniting deep neural networks with established modeling principles, both predictive accuracy and interpretability have been enhanced. For instance, \emph{physics‐informed neural networks} embed physical constraints—such as boundary conditions and conservation laws—into high‐capacity models, yielding considerable gains in data efficiency \cite{Toth2023, gao2023active, gao2024coordinate, si2025complex, si2025initialization, luoneural}; recent training/optimization refinements further improve stability and accuracy, e.g., convolution-weighting with a primal–dual perspective for PINNs \cite{si2025convolution}. In parallel, specialized approaches for PDE‐driven systems—ranging from convolutional operators to graph neural networks—have been devised to capture multi‐scale or interacting particle dynamics \cite{Freedman2022}. Recent developments, such as the Dynamic Schwartz‐Fourier Neural Operator \cite{Gao2025DSFNO}, unified spectral–physical representations for learning PDEs (Holistic Physics Solver) \cite{yue2025holisticphysicssolverlearning}, residual learning of physical trajectories for PDE solving (DeltaPhi) \cite{yue2024deltaphilearningphysicaltrajectory}, systematic analyses of discretization–mismatch errors in neural operators \cite{gao2025discretization}, and physics‐constrained generative sampling for observation-limited PDEs \cite{xu2025apod}, further expand the expressive power of operator-based techniques and continue to push the boundaries of ML-based modeling in physical sciences.

Data-driven methods for uncovering explicit governing equations have also garnered attention. Approaches like sparse regression and symbolic regression can infer interpretable representations from time-series data but often require careful regularization or domain knowledge, especially under noisy or high-dimensional conditions \cite{Xue2022}. Recent developments in deep learning have introduced physics-inspired layers and canonical coordinate transformations to bolster interpretability and maintain geometric fidelity \cite{Jacobs2021}.

Within this broader context, \emph{Hamiltonian Neural Networks (HNNs)} focus on enforcing energy conservation by learning Hamiltonians directly from data. However, when using generic multilayer perceptrons, HNNs may struggle to capture stiff or complex energy surfaces, prompting the search for robust mathematical frameworks. The Kolmogorov–Arnold Representation, which provides a powerful way to decompose high-dimensional functions into univariate building blocks, has thus become a promising tool for enhancing local approximation quality in challenging physical domains \cite{Pan2019,Tang2020}.

\section{Preliminaries} \label{sec:pre}
This section provides the mathematical groundwork for our method. Specifically, sections \ref{sec:HamiltonianDynamics} and \ref{sec:SymplecticTransformations} review key properties of Hamiltonian dynamics and symplectic transformations. Section \ref{sec:HNNs} summarizes the Hamiltonian Neural Network (HNN) framework. Section \ref{sec:KANs} introduces Kolmogorov--Arnold Networks (KANs) and highlights their relevance to Hamiltonian systems. Finally, Section \ref{sec:RelatedWork} situates our approach within the broader context of machine learning and dynamical systems research.

\subsection{Hamiltonian Dynamics} \label{sec:HamiltonianDynamics}
A \emph{Hamiltonian system} on a $2d$-dimensional manifold $\mathcal{M}$ (often $\mathbb{R}^{2d}$) is described by generalized coordinates $q \in \mathbb{R}^d$ and momenta $p \in \mathbb{R}^d$. A scalar function 
\[
H(q,p): \mathcal{M} \;\to\; \mathbb{R}
\]
governs the evolution of the system through the first-order ordinary differential equations:
\begin{equation}\label{eq:hamiltonian-system}
\dot{q} = \frac{\partial H}{\partial p}, 
\qquad
\dot{p} = -\frac{\partial H}{\partial q}.
\end{equation}
From a mathematical standpoint, these equations encode an \emph{energy-preserving} flow because the \emph{total derivative} of $H$ with respect to time vanishes along the trajectory:
\[
\frac{dH}{dt}
\;=\;
\frac{\partial H}{\partial q} \dot{q} \;+\; \frac{\partial H}{\partial p} \dot{p}
\;=\; 0.
\]
Hence, $H(q,p)$ remains constant if the system is not dissipative or subject to driving forces\cite{Marsden1999}. %\GK{"not forced or dissipative"~"not dissipative or subject to driving forces"}

\subsection{Symplectic Transformations} \label{sec:SymplecticTransformations}
A central concept in Hamiltonian mechanics is the \emph{symplectic structure}, which provides a geometric foundation for phase space. Formally, a \emph{symplectic form} $\omega$ on a manifold $\mathcal{M}$ is a closed, nondegenerate 2-form \cite{SanzSerna1994}. Being closed ($\mathrm{d}\omega = 0$) ensures consistent behavior of Hamilton’s equations across $\mathcal{M}$, while nondegeneracy guarantees that coordinates and momenta pair uniquely.

A transformation
\[
\Phi: (q, p) \,\mapsto\, (Q, P)
\]
is called \emph{symplectic} if it satisfies the pullback condition $\Phi^*\omega = \omega$ \cite{Marsden1999}. Intuitively, this means $\Phi$ preserves the “area” measured by $\omega$, retaining the canonical structure of Hamiltonian systems. Such preservation is crucial for \emph{bounded energy error} over long horizons; non-symplectic methods may exhibit accumulating energy drift \cite{Hairer2006}.

In practice, symplectic integrators like leapfrog and Verlet explicitly maintain $\Phi^*\omega = \omega$ at each step \cite{SanzSerna1994}, yielding more stable long-term trajectories. Recent advances also embed symplectic principles into \emph{data-driven} models, where neural architectures are constrained to preserve $\omega$ \cite{Greydanus2019}. This approach fuses learning from observations with robust energy and phase-space volume conservation, making it valuable for high-dimensional, long-duration simulations.

\subsection{Hamiltonian Neural Networks (HNNs)} \label{sec:HNNs}
%Hamiltonian Neural Networks \cite{ChenWang2021} aim to incorporate the geometric and energy-preserving properties of Hamiltonian flows into a neural model. Rather than predicting $(\dot{q}, \dot{p})$ directly, an HNN learns a parameterized Hamiltonian $H_\theta(q,p)$ and enforces:
%\[
%\dot{q} \;=\; \frac{\partial H_\theta(q,p)}{\partial p},
%\quad
%\dot{p} \;=\; -\frac{\partial H_\theta(q,p)}{\partial q}.
%\]
%In principle, this guarantees an energy-like conserved quantity $H_\theta$ over each training trajectory, although the approximation quality depends on how well $H_\theta$ can represent the true Hamiltonian. In many implementations, $H_\theta$ is realized via a feedforward network, but this can pose challenges when localizing features or handling high-frequency components, \LC{\sout{thus motivating other architectures.} prompting exploration of alternative architectures}

Hamiltonian Neural Networks \cite{ChenWang2021} aim to incorporate the geometric and energy-preserving properties of Hamiltonian flows into a neural model. Rather than predicting $(\dot{q}, \dot{p})$ directly, an HNN learns a parameterized Hamiltonian $H_\theta(q,p)$ and enforces:
\[
\dot{q} \;=\; \frac{\partial H_\theta(q,p)}{\partial p},
\quad
\dot{p} \;=\; -\frac{\partial H_\theta(q,p)}{\partial q}.
\]
In principle, this guarantees an energy-like conserved quantity $H_\theta$ for each training trajectory, although the fidelity of that conservation depends on how effectively $H_\theta$ approximates the true Hamiltonian. In many implementations, $H_\theta$ is realized through a feedforward network, but this can pose challenges when localizing features or handling high-frequency components, prompting exploration of alternative architectures.

\subsection{Kolmogorov--Arnold Networks (KANs)}
\label{sec:KANs}
Inspired by the Kolmogorov–Arnold Representation theorem \cite{Kolmogorov1957,Arnold1957}, \emph{Kolmogorov--Arnold Networks (KANs)} decompose a multivariate function into sums and compositions of univariate transformations. Specifically, for a continuous function 
\[
f: [0,1]^n \to \mathbb{R},
\]
this theorem guarantees an existence form:
\[
f(\mathbf{x})
\;=\;
\sum_{q=1}^{2n+1}
\Phi_q \!\Bigl(
\sum_{p=1}^{n} \phi_{q,p}(x_p)
\Bigr)
\]
where $\phi_{q,p}$ and $\Phi_q$ are continuous \emph{univariate} functions. Drawing on this fundamental result, researchers have developed KANs as neural architectures that implement these univariate blocks—often small MLPs or other basis expansions—to approximate high-dimensional mappings \cite{ZimingLiu2023,Pan2019}.
To leverage these advantages, KANs are designed with three key features:

\noindent\textbf{Localized Approximation:} Unlike MLPs, which often parameterize a function globally, KANs employ univariate modules allocated for specific components. This strategy enables the network to capture multi-scale or stiff behaviors in each variable dimension, especially useful when dealing with large or highly oscillatory domains.

\noindent\textbf{Smooth Gradients:} Since each subfunction in KANs is univariate (e.g., low-order polynomials or radial basis functions), their partial derivatives remain smooth and well-behaved. This ensures stability and numerical consistency in derivative-sensitive tasks such as Hamiltonian modeling \cite{Tang2020}. We note that smooth gradients are also particularly critical in the context of partial-update optimization \cite{xu2025impactschemessimulatedannealing}.

\noindent\textbf{Reduced Hyperparameter Tuning:} By constructing the overall function from sums and compositions of simpler blocks, KANs obviate the need for excessively deep or wide neural architectures. As a result, practitioners can avoid extensive hyperparameter searches, which is particularly advantageous for high-dimensional inputs.

KAN’s ability to locally adapt and represent complex behaviors makes it particularly suited to Hamiltonian dynamics, where energy functions $H(q,p)$ often exhibit localized or oscillatory patterns. Moreover, because the Kolmogorov–Arnold Representation remains a single scalar function once composed, plugging $H_\theta$ into Hamilton’s equations ensures that the core physical properties—such as symplectic structure and energy conservation—are preserved. This observation motivates our proposed approach, \emph{Kolmogorov--Arnold Representation-based Hamiltonian Neural Network (KAR-HNN)}.

\section{Method: Kolmogorov--Arnold Representation-Based HNN} 
\label{sec:method}

This section lays out the theoretical foundations and key methodological steps of our proposed \emph{Kolmogorov--Arnold Representation-based Hamiltonian Neural Network (KAR-HNN)}. This section clarifies how and why this approach preserves key physical properties—energy conservation and symplectic geometry—while providing a constructive method to learn high-dimensional Hamiltonian dynamics. We also highlight how the learnable parameters arise from the Kolmogorov–Arnold Representation principle.

\subsection{Model Formulation via the Kolmogorov–Arnold Representation}

We consider a Hamiltonian system in a 2d-dimensional phase space \( \mathcal{M} \subset \mathbb{R}^{2d} \), where \( q \in \mathbb{R}^d \) are generalized coordinates and \( p \in \mathbb{R}^d \) are the corresponding momenta. The time evolution follows
\[
\dot{q} = \frac{\partial H(q,p)}{\partial p},
\quad
\dot{p} = -\,\frac{\partial H(q,p)}{\partial q},
\]
with \( H: \mathcal{M} \to \mathbb{R} \) denoting the total energy. Rather than modeling \(H\) as a generic black box, we take advantage of the Kolmogorov–Arnold Representation to structure \(H_\theta\) with univariate components.

\begin{theorem}[Kolmogorov--Arnold Representation \cite{Kolmogorov1957,Arnold1957}]
Let \( f: [0,1]^n \to \mathbb{R} \) be a continuous function of n variables. There exist continuous univariate functions
\[
\phi_{r,s}: [0,1] \to \mathbb{R}
\quad
\text{and}
\quad
\Phi_{r}: \mathbb{R} \to \mathbb{R},
\]
for \(1 \le r \le 2n+1\) and \(1 \le s \le n,\) such that
\[
f(x_1, x_2, \dots, x_n)
=
\sum_{r=1}^{2n+1}
\Phi_r\biggl(
\sum_{s=1}^n \phi_{r,s}(x_s)
\biggr).
\]
\end{theorem}

To adapt this to our Hamiltonian setting, define the state vector
\[
z = [q \,;\, p] \in \mathbb{R}^{2d},
\]
and write
\[
H_\theta(z)
=
\sum_{r=1}^{2d+1}
\Phi_r^\theta
\Bigl(
\sum_{s=1}^d
\phi_{r,s}^\theta(z_s)
\Bigr),
\]
where each univariate map \(\phi_{r,s}^\theta\) or \(\Phi_r^\theta\) is parameterized by \(\theta\). Taken together, these univariate pieces form the scalar energy \(H_\theta(z)\).

Once \(H_\theta\) is defined, the Hamiltonian equations become
\[
\dot{q} = \nabla_p H_\theta(q,p),
\quad
\dot{p} = -\,\nabla_q H_\theta(q,p).
\]
We compute these partial derivatives by applying the chain rule to each univariate component, ensuring that every parameter in \(\theta\) has a direct impact on the learned energy landscape.

\subsection{Training from Observed Trajectories}

Assume we have a dataset
\[
\mathcal{D} 
= 
\bigl\{\,(q(t_j), p(t_j)),\;(\dot{q}(t_j), \dot{p}(t_j))\,\bigr\}_{j=1}^{M},
\]
sampled at times \(t_1, t_2, \dots, t_M\). Let 
\(\mathbf{z}(t_j) = [\,q(t_j);\, p(t_j)\,]\) and 
\(\dot{\mathbf{z}}(t_j) = [\,\dot{q}(t_j);\, \dot{p}(t_j)\,]\). Our goal is to align the KA-based Hamiltonian derivatives with these observations:
\[
\dot{q}(t_j) \;\approx\; \nabla_{p} H_\theta\!\bigl(q(t_j),p(t_j)\bigr),
\quad
\dot{p}(t_j) \;\approx\; -\,\nabla_{q} H_\theta\!\bigl(q(t_j),p(t_j)\bigr).
\]
We adopt a mean-squared error loss:
\[
\begin{aligned}
\mathcal{L}(\theta)
&=
\frac{1}{M}
\sum_{j=1}^{M}
\Bigl\|
\dot{q}(t_j)
\;-\;
\nabla_p H_\theta\bigl(q(t_j),p(t_j)\bigr)
\Bigr\|^2 \\[6pt]
&\quad+
\Bigl\|
\dot{p}(t_j)
\;+\;
\nabla_q H_\theta\bigl(q(t_j),p(t_j)\bigr)
\Bigr\|^2,
\end{aligned}
\]
which is minimized via a gradient-based optimizer (e.g., Adam). Automatic differentiation handles the univariate partial derivatives \(\phi_{r,s}^{(\theta)}\) and \(\Phi_r^{(\theta)}\).

\subsection{Symplectic Preservation}

Although the above derivations may appear theoretical, they explain why the learned model retains both energy and geometric fidelity. In Hamiltonian mechanics, a flow is called symplectic if it preserves the standard symplectic 2-form \(\omega_0\). Specifically, we define
\[
\omega_0\bigl((q_1,p_1),(q_2,p_2)\bigr)
=
q_1^\top p_2 - p_1^\top q_2
\]
where \(\mathbf{z} = (q,p)\). A map \(\Phi_t: \mathbb{R}^{2d}\to \mathbb{R}^{2d}\) is symplectic if \(\Phi_t^*\omega_0 = \omega_0\).

Any Hamiltonian system on \(\mathbb{R}^{2d}\) can be written as
\[
\dot{\mathbf{z}} = X_H(\mathbf{z}) = J\,\nabla_{\mathbf{z}}\,H(\mathbf{z}),
\quad
\text{where}
\quad
J=
\begin{pmatrix}
0 & I_d\\
-I_d & 0
\end{pmatrix}
\]
Hence,
\[
X_H(\mathbf{z})
=
\begin{pmatrix}
\dot{q}\\
\dot{p}
\end{pmatrix}
=
\begin{pmatrix}
\nabla_p H(\mathbf{z})\\
-\nabla_q H(\mathbf{z})
\end{pmatrix}.
\]
Any continuously differentiable scalar function \(H(\mathbf{z})\) generates a symplectic flow because \(\iota_{X_H}(\omega_0) = dH\) implies \(\mathcal{L}_{X_H}\omega_0 = 0\).

In the Kolmogorov–Arnold representation-based Hamiltonian Neural Network (KAR-HNN), the Hamiltonian \(H_\theta(\mathbf{z})\) is composed of univariate building blocks but remains a single, continuous scalar function:
\[
H_\theta(\mathbf{z})
=
\sum_{r=1}^{2d+1}
\Phi_r^\theta
\Bigl(
\sum_{s=1}^d
\phi_{r,s}^\theta(z_s)
\Bigr).
\]
Substituting \(H_\theta\) into the Hamiltonian equations ensures \(\dot{\mathbf{z}} = J\,\nabla_{\mathbf{z}}\,H_\theta(\mathbf{z})\), thereby making the induced flow symplectic. Consequently, energy surfaces remain intact and trajectory drift is reduced over extended time horizons, mirroring the behavior of classical Hamiltonian systems. These theoretical guarantees extend naturally to real-world modeling, even in high-dimensional problems.

By embedding the Kolmogorov–Arnold Representation into a Hamiltonian framework, KAR-HNN offers a principled way to learn complex energy landscapes while upholding the core structure of Hamiltonian flows. Numerical experiments in the following section demonstrate more accurate long-term predictions and stronger energy conservation than typical MLP-based HNNs.

\vspace{-2ex}
\section{Experiments}
\label{sec:Experiment}
% . All three modeling approaches—(1) a baseline network without Hamiltonian priors, (2) an MLP-based HNN, and (3) our proposed KAR-HNN—employ a derivative-matching objective.
We present four classic systems of increasing complexity—spring–mass, simple pendulum, two‐body, and three‐body. The baseline and MLP‐HNN rely on typical feedforward layers, while KAR‐HNN replaces them with univariate expansions via the Kolmogorov–Arnold Representation. 

\subsection{Spring--Mass System}
Consider a classical mass--spring system with mass \(m = 1\) and spring constant \(k = 1\). The Hamiltonian is given by
\[
H(q,p) \;=\; \frac{p^2}{2m} \;+\; \tfrac12\,k\,q^2,
\]
where \(q\) measures displacement from equilibrium, and \(p = m\,\dot{q}\) is the momentum (see Fig.~\ref{fig:spring_data} for a representative trajectory).

We generated 25 training and 25 test trajectories, each containing 30 samples, with total energies from 0.2 to 1.0 to encompass a range of oscillation amplitudes. Gaussian noise with \(\sigma^2 = 0.1\) was optionally added to reflect real-world measurement uncertainty.

Three models were evaluated: a baseline neural network, a Hamiltonian Neural Network (HNN) using a multilayer perceptron (MLP), and the proposed KAR-HNN. The baseline and MLP-HNN both contained 200 hidden units, used a learning rate of \(10^{-3}\), a weight decay of \(10^{-4}\), and were trained for 2000 iterations using Adam. In contrast, the KAR-HNN utilized a univariate hidden layer of two nodes, \(\mathit{grid} = 2\), \(k = 5\) in B-spline expansions, and was trained via LBFGS for 200 steps with a batch size of 50. These choices underscore the differences between large, fully connected networks and the localized function expansions inspired by the Kolmogorov–Arnold Representation.

Table~\ref{tab:spring} reports the training and test mean squared errors (MSE) for \((\dot{q},\dot{p})\) (scaled by \(10^3\)) as well as an energy-error metric indicating how closely each model preserves the Hamiltonian over time. The baseline achieves MSE values around 36--37 for both training and testing, but exhibits high energy drift exceeding 160, reflecting minimal constraint to a constant-energy surface. By contrast, the MLP-HNN slightly reduces the test MSE (to roughly 34--35) yet drastically lowers energy drift to about 0.4, demonstrating that a Hamiltonian prior can significantly mitigate artificial energy fluctuations. Meanwhile, KAR-HNN attains the strongest test MSE (around 28--30), more accurately capturing the phase-space derivatives. Although its energy drift (near 1.6) is larger than that of the MLP-HNN, it remains substantially lower than the baseline’s, indicating that localized expansions can flexibly model the oscillator’s behavior even if strict energy conservation is mildly compromised.

\begin{table}[t]
    \centering
    \renewcommand{\arraystretch}{1.2}
    \caption{Spring--mass system performance (scaled by $10^3$). Energy error reflects average deviation of $H$ over time.}
    \label{tab:spring}
    \begin{tabular}{@{}lccc@{}}
    \hline
    \textbf{Model} & \textbf{Train Loss} & \textbf{Test Loss} & \textbf{Energy} \\
    \hline
    Baseline & 37.1 $\pm$ 1.91 & 36.6 $\pm$ 1.86 & 168 $\pm$ 20.5 \\
    MLP-HNN  & 36.9 $\pm$ 1.91 & 35.9 $\pm$ 1.83 & 0.376 $\pm$ 0.0798 \\
    KAR-HNN  & 35.2 $\pm$ 7.70 & 28.6 $\pm$ 5.59 & 1.63  $\pm$ 0.284 \\
    \hline
    \end{tabular}
\end{table}
\FloatBarrier

\subsection{Simple Pendulum}

\begin{figure*}[!t]
  \centering
  \subfloat[Spring–mass phase portrait.\label{fig:spring_data}]{
    \begin{overpic}[width=0.45\textwidth]{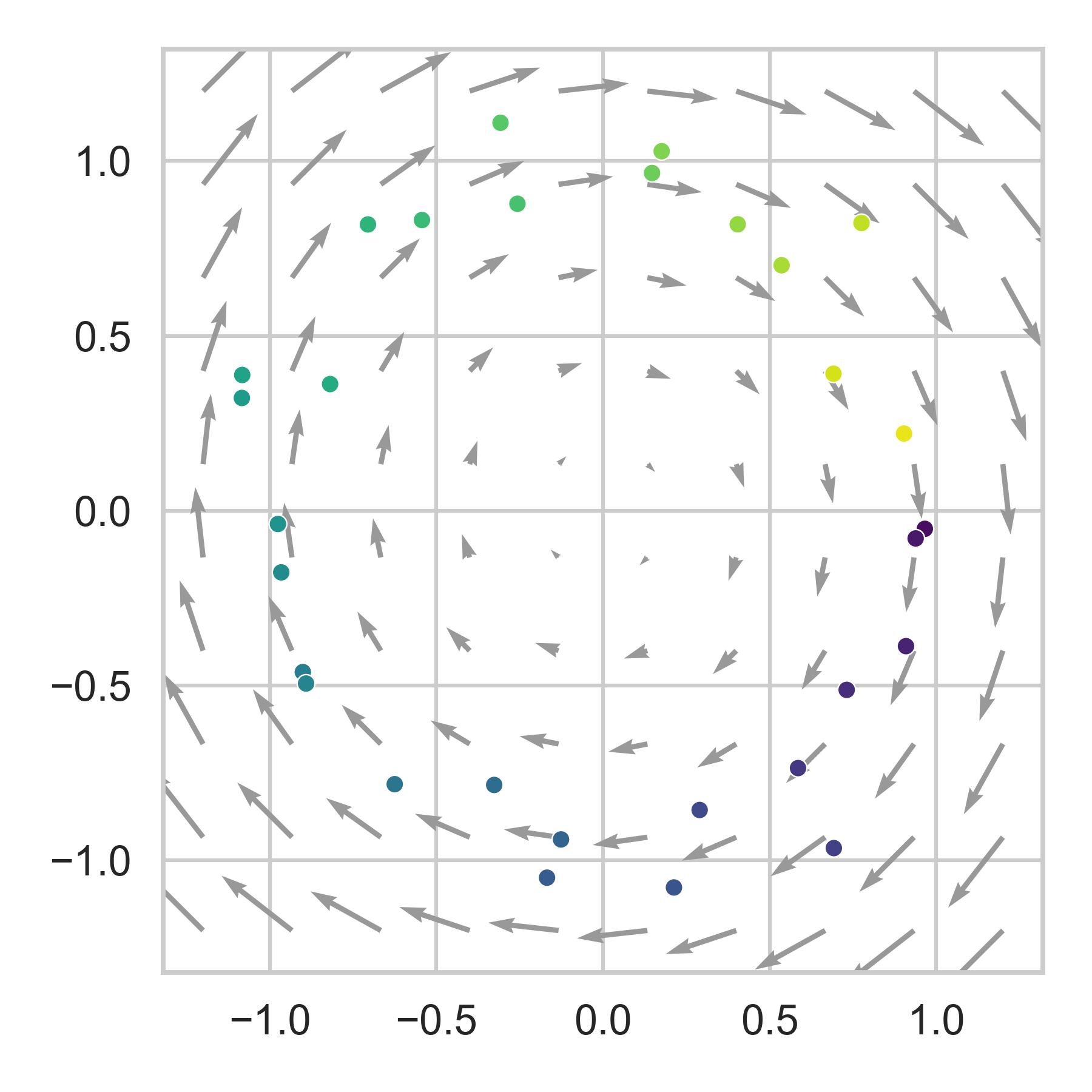}
      \put(1,53){\Large $p$}
      \put(53,0){\Large $q$}
    \end{overpic}
  }
  \hfill
  \subfloat[Simple pendulum phase portrait.\label{fig:pend_data}]{
    \begin{overpic}[width=0.45\textwidth]{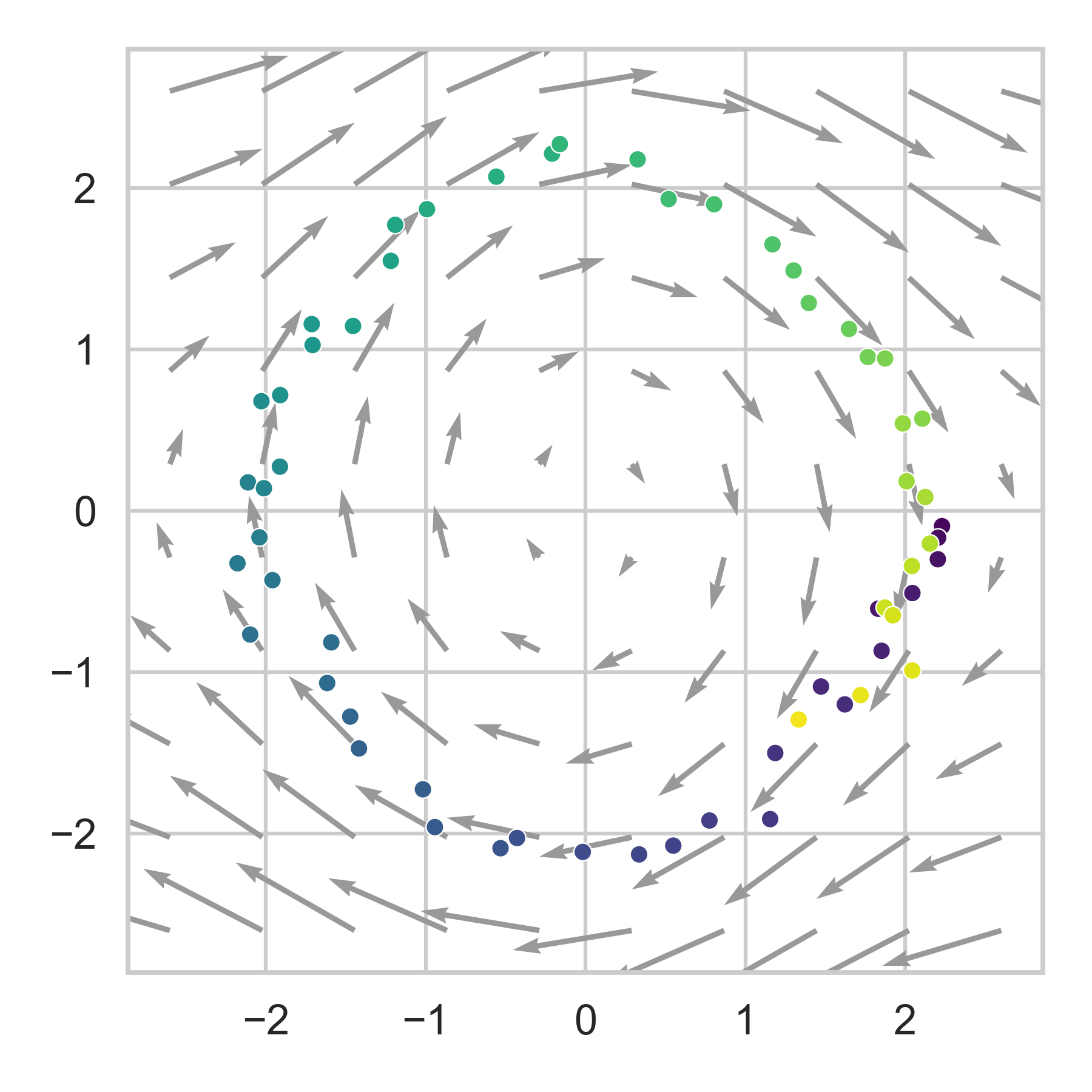}
      \put(1,53){\Large $p$}
      \put(53,0){\Large $q$}
    \end{overpic}
  }
  \caption{Datasets for two classical systems.}
  \label{fig:combined_fig}
\end{figure*}

We next consider a simple pendulum of length \(\ell = 1\) and mass \(m = \tfrac12\). Its Hamiltonian is
\[
H(q,p) \;=\; \frac{p^2}{2\,m\,\ell^2} \;+\; 2\,m g \ell \,\bigl(1 - \cos q\bigr),
\]
where \(p = m\,\ell^2\,\dot{q}\) is the momentum and \(g = 3\). Fig.~\ref{fig:pend_data} shows the training dataset, consisting of 25 training and 25 test trajectories with 45 samples each. Gaussian noise with \(\sigma^2 = 0.1\) was optionally added to simulate measurement imperfections.

As before, we compare a baseline neural network, a Hamiltonian Neural Network (HNN) with an MLP, and the proposed KAR-HNN. The baseline and MLP-HNN share hyperparameters with the spring--mass experiment, while KAR-HNN adjusts the B-spline parameter \(k\) to 3. Table~\ref{tab:pend} presents the mean squared errors (MSE) for \((\dot{q}, \dot{p})\), scaled by \(10^3\), along with an energy-error metric indicating deviations from the constant-energy surface.

The baseline model attains a training loss of about 31.3 but deteriorates on the test set to 38.0, accompanied by a sizeable energy drift of 41.2. By contrast, the HNN achieves comparable training performance (31.7) and a similar test loss (38.5), yet its energy deviation (28.8) is noticeably smaller, reflecting the benefit of embedding Hamiltonian structure. KAR-HNN obtains the best test MSE at 34.3, coupled with the lowest energy drift of 24.6 among the three methods. Although the baseline can learn some aspects of the pendulum’s motion superficially, it fails to maintain accuracy across unobserved swings, underscoring how explicit Hamiltonian constraints or localized expansions bolster long-term predictive fidelity.

\begin{table}[t]
    \centering
    \renewcommand{\arraystretch}{1.2}
    \caption{Simple pendulum results (scaled by $10^3$).}
    \begin{tabular}{@{}lccc@{}}
    \hline
    \textbf{Model} & \textbf{Train Loss} & \textbf{Test Loss} & \textbf{Energy}\\
    \hline
    Baseline & 31.3 $\pm$ 1.76 & 38.0 $\pm$ 2.13 & 41.2 $\pm$ 7.33 \\
    HNN      & 31.7 $\pm$ 1.77 & 38.5 $\pm$ 2.10 & 28.8 $\pm$ 6.94\\
    KAR-HNN  & 32.8 $\pm$ 1.76 & 34.3 $\pm$ 1.75 & 24.6 $\pm$ 6.94\\
    \hline
    \end{tabular}
    \vspace{1ex}
    \label{tab:pend}
\end{table}

% ---- Datasets for Spring & Pendulum (kept combined), placed after both subsections ----
\FloatBarrier

\subsection{Two-Body Problem}

\begin{figure}[!t]
\centering
\begin{overpic}[width=\linewidth]{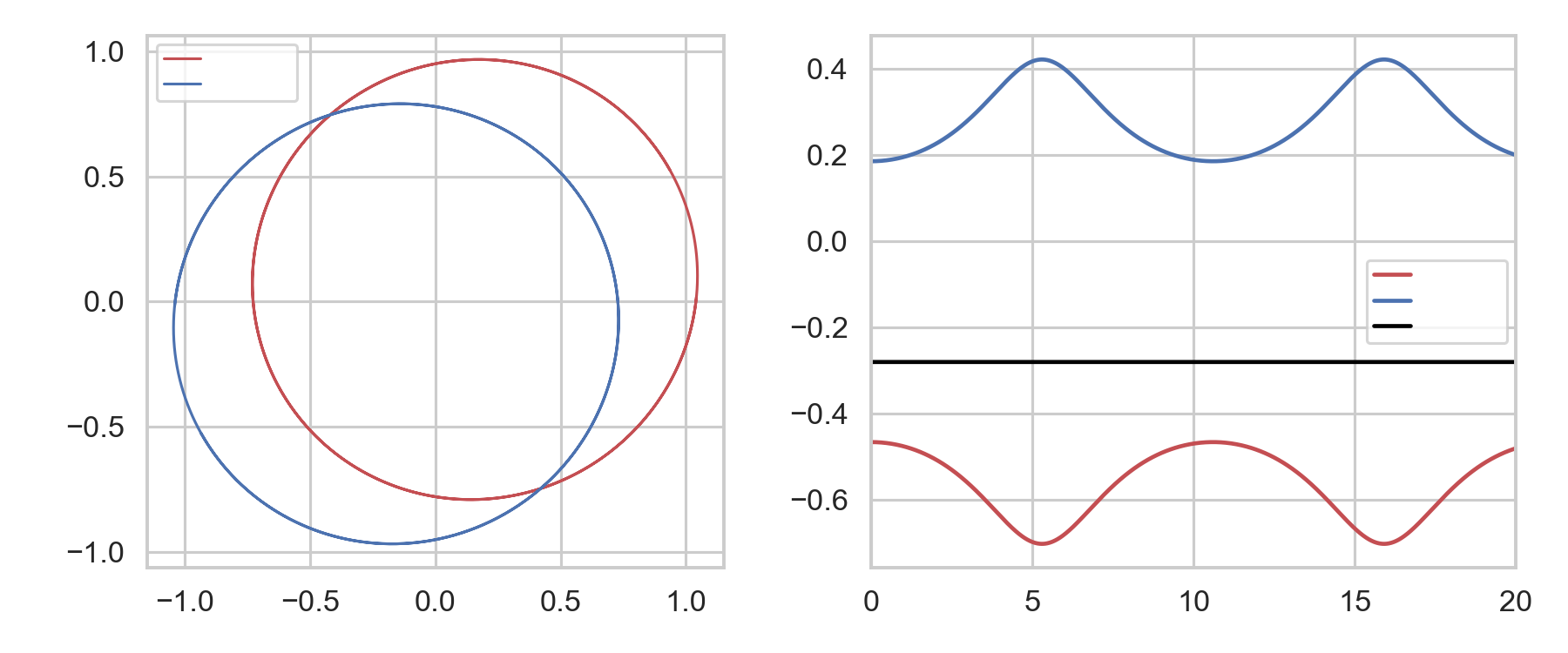}
  \put(0,20){\large $y$}
  \put(27,-1){\large $x$}
  \put(72,-1){\small time}
  \put(20,41){\small Trajectories}
  \put(69,41){\small Energy}
  \put(14,37.5){\scalebox{0.4}{body 0}}
  \put(14,36){\scalebox{0.4}{body 1}}
  \put(90.5,23.8){\scalebox{0.4}{potential}}
  \put(90.5,22){\scalebox{0.4}{kinetic}}
  \put(90.5,20.3){\scalebox{0.4}{total}}
\end{overpic}
\caption{Example trajectories for the two-body dataset (left) and the corresponding energy decomposition (right).}
\label{fig:2body-dataset}
\end{figure}
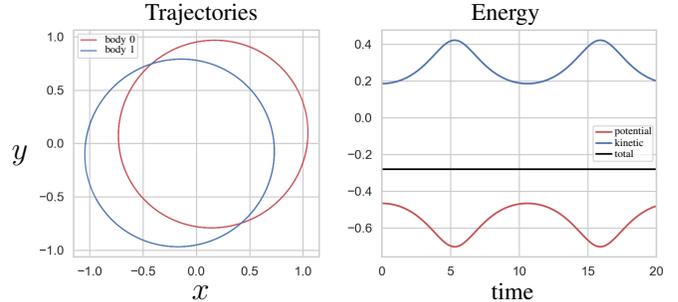

For the two-body gravitational problem, let \(\mathbf{q}\) and \(\mathbf{p}\) be the position and momentum, yielding 
\[
H(\mathbf{q},\mathbf{p})
=
\frac{\|\mathbf{p_1}\|^2+\|\mathbf{p_2}\|^2}{2\mu}
\;-\;
\frac{G\,m_1\,m_2}{\|\mathbf{q}_{1}-\mathbf{q}_{2}\|}.
\]
Fig. \ref{fig:2body-dataset} shows two masses orbiting one another under mutual gravity. The left panel plots their trajectories in the x-y plane, forming closed orbits around the system’s center of mass. The right panel shows potential energy (negative), kinetic energy (positive), and their sum (constant), illustrating energy conservation over time.We generate 1000 near-circular and elliptical orbits, each with 50 samples, noise \(\sigma^2=0.05\), and an 80:20 train–test split. The baseline and MLP-HNN each run 10000 steps (with bigger batch sizes), while KAR-HNN employs two hidden layers (10 univariate nodes each), \(grid=k=3\), 4000 steps, batch size 50.

% (keep your analysis paragraph exactly as-is)

\begin{table}[!t]
    \centering
    \renewcommand{\arraystretch}{1.2}
    \caption{Two-body results (scaled by $10^6$). }
    \begin{tabular}{@{}lccc@{}}
    \hline
    Model & Train Loss & Test Loss & Energy\\
    \hline
    Baseline & 30.3 $\pm$ 1.24 & 26.7 $\pm$ 0.786 & 1.01$e$+5 $\pm$ 2.50$e$+4 \\
    HNN      & 2.37 $\pm$ 0.166 & 1.92 $\pm$ 3.97$e$-2 & 4.99 $\pm$ 0.395 \\
    KAR-HNN  & 0.216 $\pm$ 4.03$e$-2 & 0.281 $\pm$ 4.73$e$-2 & 1.56 $\pm$ 1.18 \\
    \hline
    \end{tabular}
    \label{tab:2body}
\end{table}

% -- Energy comparison figure (full column width, top) --
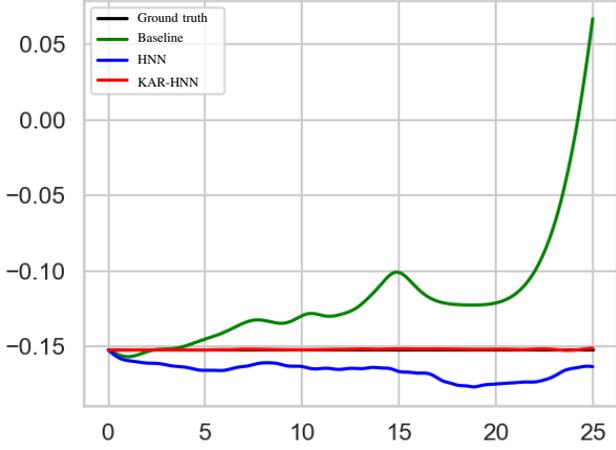
\begin{figure}[!t]
\centering
\begin{overpic}[width=\linewidth]{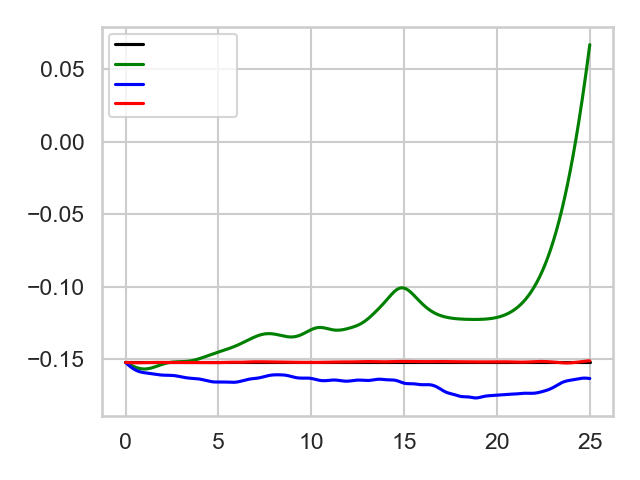}
  \put(24,67.7){\tiny Ground truth}
  \put(24,64.7){\tiny Baseline}
  \put(24,61.3){\tiny HNN}
  \put(24,58){\tiny KAR-HNN}
\end{overpic}
\caption{Comparison of total energy over time in the two-body problem. KAR-HNN closely matches the ground truth, while the baseline and MLP-HNN exhibit larger energy fluctuations.}
\label{fig:total-energy}
\end{figure}
As shown in Table~\ref{tab:2body} (scaled by $10^6$), the baseline obtains a test loss of 26.7 and extreme energy drift of 1.01$e$+5, indicating poor orbit fidelity. The MLP-HNN lowers test MSE to 1.92 and the energy to 4.99, benefiting from Hamiltonian constraints. KAR-HNN, however, pushes test MSE down to 0.281, along with an energy drift of 1.56, providing significantly more accurate trajectory prediction. These findings suggest that univariate expansions effectively approximate the gravitational potential and kinetic components, surpassing conventional feedforward networks in capturing orbital motion. The method’s local expansions thus appear critical for modeling multi-scale or varying orbital eccentricities.

\FloatBarrier

\subsection{Three-Body Problem}
% -- Three-body comparison figure placed inside this subsection --
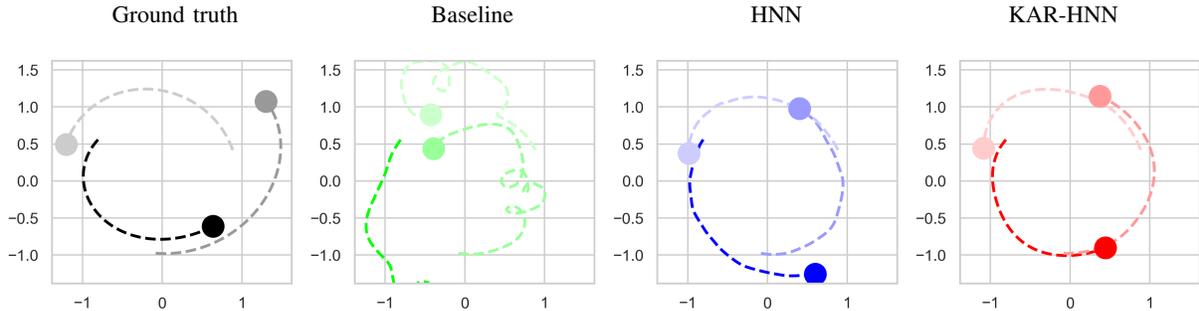
\begin{figure*}[!b]
    \centering
    \begin{overpic}[width=0.9\textwidth]{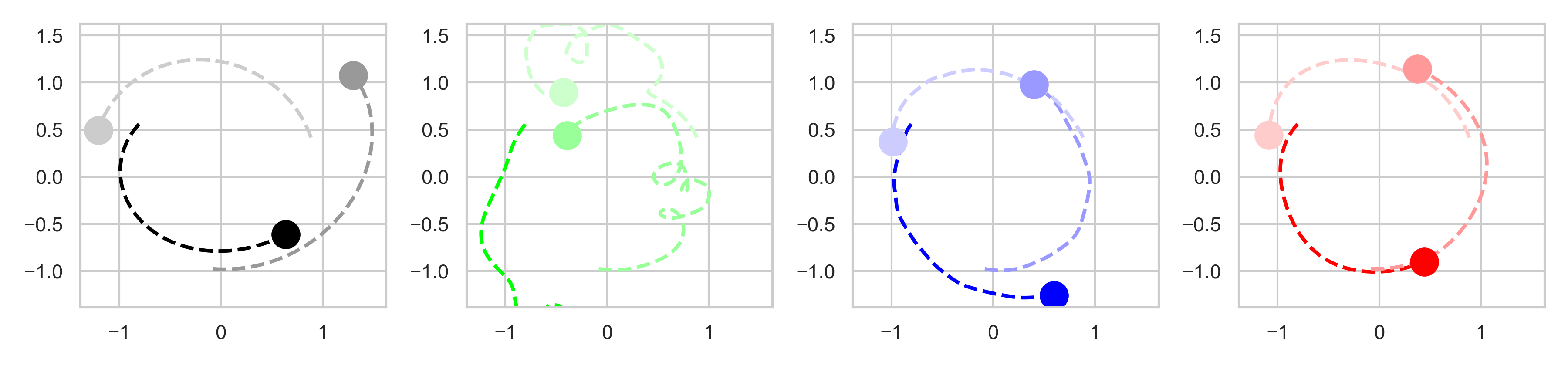}
        \put(10,25){\small Ground truth}
        \put(36,25){\small Baseline}
        \put(62,25){\small HNN}
        \put(83,25){\small KAR-HNN}
    \end{overpic}
    \vspace{-1ex}
    \caption{Sample trajectories in the three-body system. The baseline diverges early; HNN tracks partial motion but deviates significantly over time; KAR-HNN more closely follows the ground truth trajectory. This caption spans both columns.}
    \label{fig:3body-compare}
\end{figure*}
Lastly, we examine the three-body problem, with masses \((m_1,m_2,m_3)\) interacting through
\[
H
=
\sum_{i=1}^3
\frac{\|\mathbf{p}_i\|^2}{2\,m_i}
\;-\;
\sum_{1\le i<j\le3}
\frac{G\,m_i\,m_j}{\|\mathbf{q}_i-\mathbf{q}_j\|}.
\]
We collect 5000 trajectories, each with 20 observations, using radii from 0.9 to 1.2 and noise \(\sigma^2=0.2\). The baseline and MLP-HNN both train 10000 steps at batch size 600, whereas KAR-HNN uses two univariate layers (15 and 10 nodes), grid=2, $k=3$, 200 steps, batch size 50. Table~\ref{tab:3body} shows that while the MLP-HNN achieves a low energy error (2.97), it has a large test loss of 478, meaning it fails to predict chaotic motion precisely. The baseline similarly struggles, with test loss at 380. KAR-HNN, by contrast, records 17.9 in test MSE and a moderate energy drift (5.80), managing to capture chaotic orbits more accurately while still adhering to Hamiltonian principles. This highlights the method’s adaptability to high-dimensional or chaotic scenarios where small local expansions can effectively handle abrupt changes in orbital geometry.

\begin{table}[t]
    \centering
    \renewcommand{\arraystretch}{1.2}
    \caption{Three-body results (scaled by $10^3$).}
    \begin{tabular}{@{}lccc@{}}
    \hline
    Model & Train Loss & Test Loss & Energy\\
    \hline
    Baseline & 95.6 $\pm$ 65.6 & 380 $\pm$ 416 & 1.09$e$+5 $\pm$ 7.72$e$+4 \\
    HNN      & 86.3 $\pm$ 21.7 & 478 $\pm$ 450 & 2.97 $\pm$ 0.743 \\
    KAR-HNN  & 6.29 $\pm$ 2.49 & 17.9 $\pm$ 13.8 & 5.80 $\pm$ 1.54 \\
    \hline
    \end{tabular}
    \vspace{1ex}
    \label{tab:3body}
\end{table}

\subsection{Summary of Findings}

From these four systems, we observe that KAR-HNN typically achieves stronger derivative prediction (lower test loss) while preserving energy within a moderate bound, whereas the MLP-HNN often enforces near-perfect energy at the cost of higher errors in \(\dot{q},\dot{p}\). The baseline stands out for its inability to contain energy drift or generalize effectively in most tasks. Meanwhile, explicitly Hamiltonian frameworks ensure physically interpretable energy metrics, letting practitioners choose between stricter global energy adherence or greater local predictive fidelity to suit their modeling priorities.. Such outcomes suggest that local univariate expansions offer considerable advantages for multi-scale or chaotic dynamics, especially when perfect energy conservation is not mandatory. %

\clearpage  
\bibliographystyle{IEEEtran}
\bibliography{reference_Kan}

\begin{thebibliography}{10}
\providecommand{\url}[1]{#1}
\csname url@samestyle\endcsname
\providecommand{\newblock}{\relax}
\providecommand{\bibinfo}[2]{#2}
\providecommand{\BIBentrySTDinterwordspacing}{\spaceskip=0pt\relax}
\providecommand{\BIBentryALTinterwordstretchfactor}{4}
\providecommand{\BIBentryALTinterwordspacing}{\spaceskip=\fontdimen2\font plus
\BIBentryALTinterwordstretchfactor\fontdimen3\font minus \fontdimen4\font\relax}
\providecommand{\BIBforeignlanguage}[2]{{%
\expandafter\ifx\csname l@#1\endcsname\relax
\typeout{** WARNING: IEEEtran.bst: No hyphenation pattern has been}%
\typeout{** loaded for the language `#1'. Using the pattern for}%
\typeout{** the default language instead.}%
\else
\language=\csname l@#1\endcsname
\fi
#2}}
\providecommand{\BIBdecl}{\relax}
\BIBdecl

\bibitem{Hairer2006}
E.~Hairer, C.~Lubich, and G.~Wanner, \emph{Geometric Numerical Integration: Structure-Preserving Algorithms for Ordinary Differential Equations}, 2nd~ed.\hskip 1em plus 0.5em minus 0.4em\relax Springer, 2006.

\bibitem{Celledoni2021}
E.~Celledoni, H.~Z. Munthe-Kaas, B.~Owren, and G.~W. Wanner, ``On the use of canonical transformations and symplectic integrators for hamiltonian systems,'' \emph{Foundations of Computational Mathematics}, vol.~21, no.~6, pp. 1605--1638, 2021.

\bibitem{Azencot2022}
O.~Azencot, O.~Vantzos, and M.~Ovsjanikov, ``Symplectic neural networks in continuous and discrete time,'' \emph{ACM Transactions on Graphics}, vol.~41, no.~5, pp. 1--14, 2022.

\bibitem{xu2025velocityinferredhamiltonianneuralnetworks}
R.~Xu, Z.~Wu, L.~Chen, G.~Kementzidis, S.~Wang, H.~Wang, Y.~Shi, and Y.~Deng, ``Velocity-inferred hamiltonian neural networks: Learning energy-conserving dynamics from position-only data,'' \emph{arXiv preprint arXiv:2505.02321}, 2025.

\bibitem{Chen2018}
T.~Chen, Y.~Rubanova, J.~Bettencourt, and D.~K. Duvenaud, ``Neural ordinary differential equations,'' in \emph{Advances in Neural Information Processing Systems (NeurIPS)}, 2018.

\bibitem{Raissi2019}
M.~Raissi, P.~Perdikaris, and G.~E. Karniadakis, ``Physics-informed neural networks: A deep learning framework for solving forward and inverse problems involving nonlinear pdes,'' \emph{Journal of Computational Physics}, vol. 378, pp. 686--707, 2019.

\bibitem{Greydanus2019}
S.~Greydanus, M.~Dzamba, and J.~Yosinski, ``Hamiltonian neural networks,'' in \emph{Advances in Neural Information Processing Systems (NeurIPS)}, 2019.

\bibitem{Kolmogorov1957}
A.~N. Kolmogorov, ``On the representation of continuous functions of several variables by superposition of continuous functions of one variable and addition,'' \emph{Doklady Akademii Nauk SSSR}, vol. 114, pp. 953--956, 1957.

\bibitem{Arnold1957}
V.~I. Arnold, ``On functions of three variables,'' \emph{Doklady Akademii Nauk SSSR}, vol. 114, pp. 679--681, 1957.

\bibitem{Pan2019}
G.~Pan and D.~Sattigeri, ``Local kernel expansions for high-fidelity modeling of chaotic flows,'' \emph{SIAM Journal on Scientific Computing}, vol.~41, no.~6, pp. C633--C651, 2019.

\bibitem{Mishra2022}
P.~Mishra, ``Radial basis function networks in physics-informed learning,'' \emph{Applied Mathematical Modelling}, vol. 110, pp. 239--254, 2022.

\bibitem{ZimingLiu2023}
Z.~Liu, ``Kolmogorov--arnold networks for high-dimensional function approximation,'' \emph{Journal of Computational Methods}, 2023, (forthcoming).

\bibitem{Tang2020}
J.~Tang, D.~DeVito, and F.~Ebert, ``Bayesian kernelized learning for physics discovery,'' in \emph{Conference on Uncertainty in Artificial Intelligence (UAI)}, 2020.

\bibitem{Marsden1999}
J.~E. Marsden and T.~S. Ratiu, \emph{Introduction to Mechanics and Symmetry}, 2nd~ed.\hskip 1em plus 0.5em minus 0.4em\relax Springer, 1999.

\bibitem{SanzSerna1994}
J.~M. Sanz-Serna and M.~P. Calvo, \emph{Numerical Hamiltonian Problems}.\hskip 1em plus 0.5em minus 0.4em\relax Chapman and Hall, 1994.

\bibitem{Toth2023}
G.~Toth, ``Survey of physics-informed neural networks for scientific computing,'' \emph{Applied Computing and Informatics}, 2023, (in press).

\bibitem{gao2023active}
W.~Gao and C.~Wang, ``Active learning based sampling for high-dimensional nonlinear partial differential equations,'' \emph{Journal of Computational Physics}, vol. 475, p. 111848, 2023.

\bibitem{gao2024coordinate}
W.~Gao, R.~Xu, H.~Wang, and Y.~Liu, ``Coordinate transform fourier neural operators for symmetries in physical modelings,'' \emph{Transactions on Machine Learning Research}, 2024.

\bibitem{si2025complex}
C.~Si, M.~Yan, X.~Li, and Z.~Xia, ``Complex physics-informed neural network,'' \emph{arXiv preprint arXiv:2502.04917}, 2025.

\bibitem{si2025initialization}
C.~Si and M.~Yan, ``Initialization-enhanced physics-informed neural network with domain decomposition (idpinn),'' \emph{Journal of Computational Physics}, vol. 530, p. 113914, 2025.

\bibitem{luoneural}
J.~Luo, J.~Wang, H.~Wang, Z.~Geng, H.~Chen, Y.~Kuang \emph{et~al.}, ``Neural krylov iteration for accelerating linear system solving,'' in \emph{The Thirty-eighth Annual Conference on Neural Information Processing Systems}, 2025.

\bibitem{si2025convolution}
C.~Si and M.~Yan, ``Convolution-weighting method for the physics-informed neural network: A primal-dual optimization perspective,'' \emph{arXiv preprint arXiv:2506.19805}, 2025.

\bibitem{Freedman2022}
B.~Freedman, K.~Sutherland, and M.~Mueller, ``Graph neural operators for multi-scale physical simulations,'' \emph{IEEE Transactions on Neural Networks and Learning Systems}, 2022.

\bibitem{Gao2025DSFNO}
\BIBentryALTinterwordspacing
W.~Gao, J.~Luo, R.~Xu, and Y.~Liu, ``Dynamic schwartz-fourier neural operator for enhanced expressive power,'' \emph{Transactions on Machine Learning Research}, 2025. [Online]. Available: \url{https://openreview.net/forum?id=B0E2yjrNb8}
\BIBentrySTDinterwordspacing

\bibitem{yue2025holisticphysicssolverlearning}
\BIBentryALTinterwordspacing
X.~Yue, Y.~Yang, and L.~Zhu, ``Holistic physics solver: Learning pdes in a unified spectral-physical space,'' 2025. [Online]. Available: \url{https://arxiv.org/abs/2410.11382}
\BIBentrySTDinterwordspacing

\bibitem{yue2024deltaphilearningphysicaltrajectory}
\BIBentryALTinterwordspacing
X.~Yue, L.~Zhu, and Y.~Yang, ``Deltaphi: Learning physical trajectory residual for pde solving,'' 2024. [Online]. Available: \url{https://arxiv.org/abs/2406.09795}
\BIBentrySTDinterwordspacing

\bibitem{gao2025discretization}
W.~Gao, R.~Xu, Y.~Deng, and Y.~Liu, ``Discretization-invariance? on the discretization mismatch errors in neural operators,'' in \emph{The Thirteenth International Conference on Learning Representations}, 2025.

\bibitem{xu2025apod}
\BIBentryALTinterwordspacing
R.~Xu, H.~Wang, G.~Kementzidis, C.~Si, and Y.~Deng, ``{APOD}: Adaptive {PDE}-observation diffusion for physics-constrained sampling,'' in \emph{ICML 2025 Workshop on Assessing World Models}, 2025. [Online]. Available: \url{https://openreview.net/forum?id=Z1J7LJGDxH}
\BIBentrySTDinterwordspacing

\bibitem{Xue2022}
Y.~Xue, Q.~Lu, and B.~Gao, ``Symbolic learning of dynamical systems using mixed integer programming and neural approximations,'' \emph{Neural Computation}, vol.~34, no.~11, pp. 2381--2405, 2022.

\bibitem{Jacobs2021}
C.~Jacobs and E.~Bradley, ``Deep coordinate descent: Training neural networks with lagrangian mechanics,'' in \emph{International Conference on Learning Representations (ICLR)}, 2021.

\bibitem{ChenWang2021}
X.~Chen and S.~Wang, ``Improved hamiltonian neural networks with extended energy conservation,'' in \emph{Proceedings of the 28th International Conference on Neural Information Processing}, 2021.

\bibitem{xu2025impactschemessimulatedannealing}
R.~Xu, H.~Wang, and Y.~Deng, ``The impact of move schemes on simulated annealing performance,'' \emph{arXiv preprint arXiv:2504.17949}, 2025.

\end{thebibliography}

\end{document}